\documentclass[a4paper]{article}
\usepackage{fullpage}
\usepackage{hyperref}
\usepackage{multicol}
\usepackage{graphicx}
\usepackage{{amsmath,amsfonts,amssymb}}
\usepackage{epstopdf}
\usepackage{subfig}
\usepackage{todonotes}
\usepackage{multirow}
\usepackage{colortbl}
\usepackage{supertabular}
\usepackage{authblk}

\newcommand{\mbs}[1]{\ensuremath{\boldsymbol{#1}}}

\newcommand{\ex}{\mathbb{E}}

\newcommand{\thetav}{\mbs{\theta}}

\newcommand{\kv}{\mathbf{k}}

\newcommand{\x}{\mathbf{x}}
\newcommand{\y}{\mathbf{y}}

\newcommand{\K}{\mathbf{K}}

\newcommand{\X}{\mathbf{X}}

\newcommand{\w}{\mathbf{\omega}}

\newcommand{\xq}{x_{q}}

\newcommand{\N}{\mathcal{N}}

\title{%\LARGE \bf 
Unscented Bayesian Optimization for Safe Robot Grasping
}

\author[1]{Jos\'e Nogueira}
\author[2]{Ruben Martinez-Cantin}
\author[1]{Alexandre Bernardino}
\author[1]{Lorenzo Jamone}%<-this % stops a space
\affil[1]{Institute for Systems and Robotics, Instituto Superior Tecnico, Lisboa, Portugal}
\affil[2]{Centro Universitario de la Defensa, Zaragoza, Spain, 
       email: \texttt{rmcantin@unizar.es}}%
\date{}

\begin{document}

%\author{\authorblockN{Michael Shell}
%\authorblockA{School of Electrical and\\Computer Engineering\\
%Georgia Institute of Technology\\
%Atlanta, Georgia 30332--0250\\
%Email: mshell@ece.gatech.edu}
%\and
%\authorblockA{Twentieth Century Fox\\
%\and
%\authorblockN{James Kirk\\ and Montgomery Scott}
%\authorblockA{Starfleet Academy\\
%San Francisco, California 96678-2391\\

% avoiding spaces at the end of the author lines is not a problem with
% conference papers because we don't use \thanks or \IEEEmembership

%Homer Simpson\authorrefmark{2},
%Montgomery Scott\authorrefmark{3} and
%Eldon Tyrell\authorrefmark{4}}
%Telephone: (800) 555--1212, Fax: (888) 555--1212}
%\authorblockA{\authorrefmark{4}Tyrell Inc., 123 Replicant Street, Los Angeles, California 90210--4321}}

\maketitle
\thispagestyle{empty}
\pagestyle{empty}

\begin{abstract}
We address the robot grasp optimization problem of unknown objects considering uncertainty in the input space. Grasping unknown objects can be achieved by using a trial and error exploration strategy. Bayesian optimization is a sample efficient optimization algorithm that is especially suitable for this setups as it actively reduces the number of trials for learning about the function to optimize. In fact, this active object exploration is the same strategy that infants do to learn optimal grasps. One problem that arises while learning grasping policies is that some configurations of grasp parameters may be very sensitive to error in the relative pose between the object and robot end-effector. We call these configurations unsafe because small errors during grasp execution may turn good grasps into bad grasps. Therefore, to reduce the risk of grasp failure, grasps should be planned in safe areas.  We propose a new algorithm, Unscented Bayesian optimization that is able to perform sample efficient optimization while taking into consideration input noise to find safe optima. The contribution of Unscented Bayesian optimization is twofold as if provides a new decision process that drives exploration to safe regions and a new selection procedure that chooses the optimal in terms of its safety without extra analysis or computational cost. Both contributions are rooted on the strong theory behind the unscented transformation, a popular nonlinear approximation method. We show its advantages with respect to the classical Bayesian optimization both in synthetic problems and in realistic robot grasp simulations. The results highlights that our method achieves optimal and robust grasping policies after few trials while the selected grasps remain in safe regions.
\end{abstract}

%\IEEEpeerreviewmaketitle

%

\section{Introduction}

Robot grasping in general conditions is a challenging problem due to a multitude of sources of uncertainty. Perception of object shape is often noisy and partial; geometric transformations between the perceptual reference frame and the robot end-effector may carry calibration errors. Also, link flexibility and controller errors may prevent a precise positioning of the robot hand in the desired pose for the object to grasp. As such, when planning for a robotic grasp, one must consider the impact of the uncertainty in the input space (positioning errors) in the quality of the grasp. This depends not only on the magnitude of the errors but also on the characteristics of the object and the type of grasp strategy employed. Whereas, for simple grippers, we may reason that smooth object surfaces will present less sensitivity to errors, for multi-fingered hands this is not sufficient. This problem is amplified by the noisy and discontinuous nature of commonly used grasp quality metrics. In general, a sampling approach may be required to search for good grasps in a trial and error fashion.

The question we address in the paper is: given an unknown object, where should the robot perform a grasp, in order to simultaneously maximize the grasp quality and minimize the risk of failure? A brute-force approach would need to test grasps in many different configurations in search for the best grasping point and, for each configuration, repeat the test many times to average out the robot positioning uncertainty. This is clearly unfeasible in practice, and better search strategies must be devised. In robotics in general and, for the problem of robot grasping in particular, Bayesian optimization has been one of the most successful trial-and-error techniques \cite{Kroemer2009,Kroemer2010,boularias2014efficient,ActiveRewardLearning}, even in the presence of mechanical failures \cite{Cully2015}. Bayesian optimization \cite{mockus1994,jones1998,brochu2010a} is a global optimization technique for black-box functions. Because it is designed for sample efficiency, at the cost of extra computations, it is intended for functions that are expensive to evaluate in terms of cost, energy, time, etc.

The beauty of Bayesian optimization is its capability to deal with general black-box functions, therefore being able to address the grasping problem without any extra information, just the results from previous trials. Bayesian optimization relies on a probabilistic surrogate function (e.g.: a Gaussian process) that is able to learn about the target function based on previous samples and, therefore, drive future sampling more efficiently. Because Bayesian optimization is purely driven by sampling, but, at the same time, it is currently the most sample efficient global optimization method, it has been called the \emph{intelligent brute-force} algorithm. 

However, to the authors knowledge, the consideration of uncertainty in the input space has been addressed neither in the grasp planning literature nor in the Bayesian optimization literature. The main contribution of the paper addresses the problem of input noise in Bayesian optimization, which is then applied to robot grasping. For dealing the input noise, we need a system to propagate the noise distribution from the input query through all the functions and surrogates of our method. We solve this with the unscented transformation \cite{julier2004,vdMerwe04Thesis}, a method to estimate the results of applying a nonlinear transformation to a probability distribution. Thus, we can apply the unscented transformation to propagate the input noise through all the Bayesian optimization procedure. The unscented transformation is very popular in the control and signal processing literature for its efficiency and robustness, especially when combined with the Kalman filter or related algorithms.

In this paper, we present the Unscented Bayesian optimization (UBO) algorithm. It has the advantages of the intelligent brute-force from Bayesian optimization and the capability of dealing with input noise during function queries. Applied to grasping, this means that the method can find the optimal grasp while considering the input noise for safety. Furthermore, due to the recent popularity of Bayesian optimization in many areas, this method can be automatically extended to many other fields. Recent examples are applications in autonomous algorithm tuning \cite{Snoek2012,feurer2015efficient}, robot planning \cite{marchant2014bayesian,MartinezCantin09AR}, control \cite{Tesch_2011_7370,Calandra2015a}, reinforcement learning \cite{MartinezCantin07RSS,ActiveRewardLearning}, product design \cite{taddy2009bayesian,forrester2006optimization}, sensor networks \cite{Srinivas10,Garnett2010}, simulation design \cite{brochu2010a}, biology \cite{ulmasov2016bayesian}, etc. All those fields could benefit for an extension to deal with input noise.

\section{Bayesian optimization}
\label{sec:bo}
Consider the problem of finding the minimum of an unknown real valued function $f:\mathbb{X} \rightarrow \mathbb{R}$, where $\mathbb{X}$ is a compact space, $\mathbb{X} \subset \mathbb{R}^d, d \geq 1$. In order to find the minimum, the algorithm has a maximum budget of $N$ evaluations of the target function $f$. The purpose of the Bayesian optimization algorithm is to select the best query points at each iteration such as the optimization gap $|y^* - y_n|$ is minimum for the available budget.

Bayesian optimization achieves optimal sampling performance by using two ingredients. First, a probabilistic surrogate model that is a distribution over the family of functions $P(f)$ where the target function $f()$ belongs. This surrogate model is able to capture all information available, from prior information to any sample or observation as soon as it is gathered. Second, an Bayesian decision process that takes all the information captured in the surrogate model and selects the next query point in order to maximize the information about the maximum. In that way, Bayesian optimization can be understood as active learning applied to learn the optimum location.

Without loss of generality, for the remainder of the paper we are going to assume that the surrogate model $P(f)$ is a Gaussian process $\mathcal{GP}(\x|\mu,\sigma^2,\thetav)$ with inputs $\x \in \mathbb{X}$, scalar outputs $y \in \mathbb{R}$ and an associated kernel or covariance function $k(\cdot,\cdot)$ with hyperparameters $\thetav$. The hyperparameters are estimated using a Monte Carlo Markov Chain (MCMC) algorithm, i.e.: slice sampling \cite{Snoek2012,MartinezCantin14jmlr}, resulting in $m$ samples $\mathbf{\Theta} = \{\thetav_i\}_{i=1}^m$.

Given a dataset at step $n$ of query points $\X = \{\x_{1:n}\}$ and its respective outcomes $\y = \{y_{1:n}\}$, then the prediction of the Gaussian process at a new query point $\x_q$, with kernel $k_i$ conditioned on the $i$-th hyperparameter sample $k_i = k(\cdot,\cdot|\thetav_i)$ is a normal distribution such as $y_q \sim \sum_{i=1}^m \N(\mu_i,\sigma^2_i|\x_q)$ where:
\begin{equation}
  \begin{split}
\label{eq:predgp}
\mu_i(\x_q) &= \kv_i(\x_q,\X) \K_i(\X,\X )^{-1} \y  \\
\sigma^2_i(\x_q) &= k_i(\x_q,\x_q) - \kv_i(\x_q,\X) \K_i(\X,\X )^{-1} \kv_i(\X,\x_q)    
  \end{split}
\end{equation}
being $\kv_i(\x_q,\X)$ the corresponding cross-correlation vector of the query point $\x_q$ with respect to the dataset $\X$ and $\K_i(\X,\X)$ is the Gram matrix corresponding to kernel $k_i$ for the dataset $\X$ included the noise term $\sigma^2_n$. The noise term is used to represent the observation noise in stochastic functions \cite{Huang06} or the nugget term for surrogate missmodeling \cite{Gramacy2012}. Note that, because we use a sampling distribution of $\theta$ the predictive distribution at any point $\x$ is a mixture of Gaussians.

To select the next point at each iteration, we use the \emph{expected improvement} criterion \cite{Mockus1989} as a way to minimize the optimality gap. The expected improvement is defined as the expectation of the improvement function $I(\x) = \max(0, y_{best} - f(\x))$, where $y_{best}$ is the \emph{best outcome} until that iteration. Taking the expectation over the mixture of Gaussians of the predictive distribution, we can compute the expected improvement as:
\begin{equation}
  \label{eq:eigen}
  \begin{split}
  EI(\x) &= \ex_{p(y|\x,\thetav)} \left[\max(0,\rho - f(\x))\right] \\    
         &= \sum_{i=1}^m \left[\left(y_{best} - \mu_i\right) \Phi(z_i) + \sigma_i \phi(z_i)\right]
  \end{split}
\end{equation}
where $\phi$ and $\Phi$ are the corresponding Gaussian probability density function (PDF) and cumulative density function (CDF), being $z_i = (\rho - \mu_i)/\sigma_i$. In this case, $(\mu_i,\sigma^2_i)$ is the prediction computed with Equation (\ref{eq:predgp}).

Finally, in order to avoid bias and guarantee global optimality, we rely on an initial design of $p$ points based on \emph{Latin Hypercube Sampling} (LHS) following the recommendation in the literature \cite{Jones:1998,Bull2011,kandasamy2015high}. 
%Algorithm \ref{al:bo} summarizes the basic steps in Bayesian optimization.
%\input{tgp.tex}
\section{Unscented Bayesian optimization}

% This should go in the intro
There has been previous works that consider input noise in Gaussian processes \cite{mchutchon2011}, however, those methods propagate the input noise to the output space, which may result in an over exploration of the space during optimization.
%%%

In this paper, we propose to consider the input noise during the decision process to explore and select the regions that are safe. That is, the regions that guarantee good results even if the experiment/trial is repeated several times. Our contribution is twofold: we present the \emph{unscented expected improvement} and the \emph{unscented optimum incumbent}. Both methods are based on the \emph{unscented transformation} \cite{merwe2000,julier2004} that was initially developed for tracking and filtering applications.

\subsection{Unscented transformation}
The unscented transformation is a method to propagate probability distributions through nonlinear transformations with a trade off of computational cost vs accuracy. It is based on the principle that it is easier to approximate a probability distribution than to approximate an arbitrary nonlinear function. The unscented transformation uses a set of deterministally selected samples from the original distribution (called \emph{sigma points}) and transform them through the nonlinear function $f(\cdot)$. Then, the transformed distribution is computed based on the weighted combination of the transformed sigma points.

\begin{figure}
  \centering
  \includegraphics[width=3in]{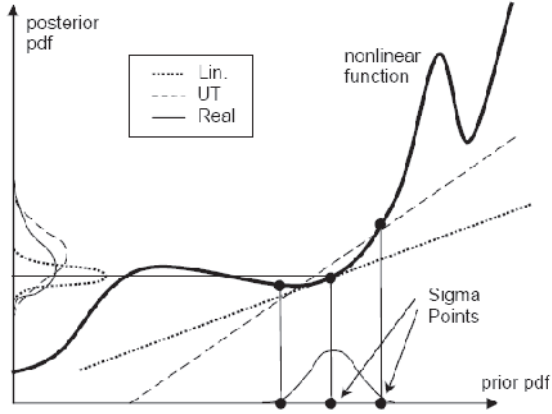}
  \caption{Propagation of a normal distribution through a nonlinear function.
The first order Taylor expansion (dotted) only uses information of the
function at the mean point to compute the linear approximation, while the UT
(dashed) approaches the function with a linear regression of several sigma
points. The actual distribution is the solid one. (Adapted from
\cite{vdMerwe04Thesis})}
  \label{fig:regression}
\end{figure}

The advantage of the unscented transformation is that the mean and covariance estimates of the new distribution are accurate to the third order of the Taylor series expansions of $f(\cdot)$ provided that the original distribution is a Gaussian prior, or up to the second order of the expansion for any other prior. Figure \ref{fig:regression} highlights the differences between approximating the distribution using sigma points (UT) or using standard first-order Taylor linearization (Lin.). The distribution from the UT is closer to the real distribution. Because the prior and posterior distributions are both Gaussians, the unscented transformation is a linearization method. However, because the linearization is based on the statistics of the distribution, it is often found in the literature as \emph{statistical linearization}.

Another advantage of the unscented transformation is its computational cost. For a $d$-dimensional input space, the unscented transformation requires a set of $2d+1$ sigma points. Thus, the computational cost is negligible compared to other alternatives to distribution approximation such as Monte Carlo, which requires a large number of samples, or numerical integration such as \emph{Gaussian quadrature}, which has an exponential cost on $d$. Van der Merwe \cite{vdMerwe04Thesis} proved that the unscented transformation is part of the more general \emph{sigma point filters}, which achieve similar performance results. Other sigma point methods are the \emph{central difference filter} (CDF) \cite{Ito00TAC} and the \emph{divided difference filter} (DDF) \cite{Norgaard00Auto}.

\subsection{Computing the unscented transformation}
Assuming that the prior distribution is a Gaussian distribution $\x \sim \N(\widehat{\x},\Sigma_\x)$, then the $2d+1$ sigma points of the unscented transformation are computed based on this sampling strategy
\begin{equation}
\begin{split}
\x^0 &= \widehat{\x}\\
\x^{(i)}_+ &= \widehat{\x} + \left(\sqrt{(d+k)\Sigma_\x}\right)_i \qquad \forall i = 1..d\\
\x^{(i)}_- &= \widehat{\x} - \left(\sqrt{(d+k)\Sigma_\x}\right)_i \qquad \forall i = 1..d
\end{split}
\label{eq:sigmapoints}  
\end{equation}
where $(\sqrt{\cdot})_i$ is the i-th row or column of the corresponding matrix square root. In this case, $k$ is a free parameter that can be used to tune the scale of the sigma points. Although it may break the positive defined requirement, the original authors \cite{julier2004} recommended $k=-3$ or $k = 0$. To alleviate the potential numerical problem raised by negative $k$ values and to increase the expressiveness of the methods, the authors later introduced the \emph{scaled unscented transform} \cite{Julier02ACC}. However, for our application such extra complexity is unnecessary. 

For these sigma points, the weights are defined as:
\begin{equation}
\begin{split}
\w^0 &= \frac{k}{d+k}\\
\w^{(i)}_+ &= \frac{1}{2(d+k)} \qquad \forall i = 1..d\\
\w^{(i)}_- &= \frac{1}{2(d+k)} \qquad \forall i = 1..d
\end{split}
\label{eq:wsigmapoints}  
\end{equation}

Then, the transformed distribution is computed as $\x' \sim \N(\widehat{\x'},\Sigma'_\x)$ where:
\begin{equation}
  \label{eq:posterior}
  \widehat{\x'} = \sum_{i=0}^{2d} \omega^{(i)} f(\x^{(i)})
  %%%% $\Sigma'_\x$
\end{equation}

\subsection{Unscented expected improvement}
Bayesian optimization is about selecting the most interesting point each iteration. Usually, this is achieve by a greedy criterion, or acquisition functions, such as: the \emph{expected improvement}, the \emph{upper confidence bound} or the \emph{predictive entropy}. Those criteria are designed to select the point that has the higher potential to become the optimum. However, all those methods assume that the observed value would be exactly the outcome of the query plus some observation noise. They assume that the query is always deterministic. 

Instead, we are going to assume that the query is a probability distribution. Thus, instead of analysing the outcome of the criterion, we are going to analyse the resulting posterior distribution of transforming the query distribution through the acquisition function. For the remainder of the section, we assume that the input distribution corresponds to the input noise in each query point $\xq$ of the Bayesian optimization process. That is, each query point is distributed according to an isotropic multivariate normal distribution $\N(0,\mathbf{I}\sigma_{x})$. 

For the purpose of safe Bayesian optimization, we will use the expected value of the transformed distribution as the acquisition function. In this case, we will use the expected improvement. Therefore, the unscented expected improvement is computed as:
\begin{equation}
  \label{eq:uei}
  UEI(\x) = \sum_{i=0}^{2d} \omega^{(i)} EI(\x^{(i)})
\end{equation}
where $\x^{(i)}$ and $\omega^{(i)}$ are computed according to equations \eqref{eq:sigmapoints} and \eqref{eq:wsigmapoints} respectively. Note that we only compute the expected value of the transformed distribution $\widehat{\x'}=UEI(\x)$. This value is enough to take a decision considering the risk on the input noise. However, the value of $\Sigma'_\x$ represents also the output uncertainty and can be used as meta-analysis tool, that is, the value can be used as a risk on the estimation of the risk. This opens a new set of criteria functions, like the upper bound of the output distribution.

\subsection{Unscented optimal incumbent}
\label{sec:unscented_optimal_selection}
The unscented expected improvement can be used to drive the search procedure towards safe regions. However, because the target function is unknown by definition, the sampling procedure can still query good outcomes in unsafe areas. 

Furthermore, in Bayesian optimization there is a final decision that is independent of the acquisition function employed. Once the optimization process is stopped after sampling $N$ queries, we still need to decide which point is the \emph{best}. Moreover, after every iterations, we may need to say which point is the incumbent as the \emph{best observation}.

If the final decision about the incumbet is based on the greedy policy of selecting the sample with best outcome $\x^*$ such that $y_{best} = f(\x^*)$ we may select an unsafe query.

Instead, we propose to apply the unscented transformation also to the select the optimal incumbent $\x^*$, based on the function outcome $f()$. This would require to evaluate on $f()$ the $2d+1$ sigma points for each observation. However, the main idea of Bayesian optimization is to reduce the number of evaluations on $f()$. Therefore, we evaluate the sigma points at the GP prediction $\mu()$. Thus, let us define the unscented outcome (UO) as:
\begin{equation}
  \label{eq:uo}
  UO(\x) = \sum_{i=0}^{2d} \omega^{(i)} \sum_{j=1}^{m} \mu_j(\x^{(i)})
\end{equation}
where $\sum_{j=1}^{m} \mu_j(\x^{(i)})$ is the prediction of the GP according to equation \eqref{eq:predgp} integrated over the kernel hyperparameters and at the sigma points of equation \eqref{eq:sigmapoints}.

Under this conditions, the incumbent of the optimal solution $\x^*$ corresponds to:
\begin{equation}
\x^* = \arg \max_{\x} \; UO(\x)
\label{eq:optuo}  
\end{equation}
In the Bayesian optimization literature, when $f()$ represents an stochastic function with large output noise, it is common to return the expected value of the GP at the optimum query, instead of the optimum observation. Note that our method is also valid under those conditions.

As an illustrative example, take the function in Fig. \ref{fig:rkhs_function}. In this case, the maximum of the function is at $x \approx 0.87$. However, this maximum is very risky, that is, small variations in $x$ results in large deviations from the optimal outcome. On the other hand, the local maximum at $x \approx 0.07$ is safer. Even if there is noise in $x$, repeated queries will produce similar outcomes. In this case, if we assume input noise of $\sigma_x = 0.05$ and compute the unscented transformation of that noise through the function, we can see that the sigma points centered at the leftmost maximum would have higher outcome that the sigma points centered at the global maximum. Therefore, the expected posterior value of the local smooth maximum would be larger than the value at the global narrow maximum.

In summary, our method takes the unscented transformation to compute the decision functions in Bayesian optimization, assuming that each query is a probability distribution (due to the input noise) instead of a deterministic value. We found that, for Bayesian optimization, we need to consider the unscented version of the acquisition function, for which we propose the unscented expected improvement. Furthermore, we also need to take into consideration the decision to select the best observation or the potential optimum. In this case, we propose the unscented optimal incumbent as a robust selection method.
\section{Results}

In this section we describe the methods used and experiments performed to compare the benefits of the Unscented Bayesian Optimization (UBO) with respect to the classical expected Bayesian optimization (BO). The main goal in these experiments is to demonstrate that, by using the UBO, we minimize the risk of choosing unsafe global optima. We first illustrate the method in synthetic functions that allows to visualize the importance of selecting the safe optimum. Then, we show the results of doing autonomous exploration of daily life objects with a dexterous robot hand using realistic simulations, reproducing the conditions of a real robot setup.

In this work we have used and extended the BayesOpt software \cite{MartinezCantin14jmlr} with the proposed methods\footnote{The new code will be release upon acceptance of the paper.}. For the kernel, we used the standard choice of the Mat{\'e}rn kernel with $\nu=5/2$. As commented in Section \ref{sec:bo}, we used slice sampling for the kernel hyperparameters.  

To reproduce the effect of the input noise, we queried the result of each method using Monte Carlo samples according the input noise distribution at each iteration. By analysing the  outcome of the samples we can estimate the expected outcome from the optima ($y_{mean} \left( \mathbf{x}_i^{mc} \right)$) and the instability of the optimum ($y_{std}\left( \mathbf{x}_i^{mc} \right)$). As we can see in the results, out method is able to provide equal or better expected outcomes while reducing the instability of those same outcomes.

    \subsection{Synthetic Functions}
    
\begin{figure}[!htb]
	\centering
	\includegraphics[width=0.45\linewidth]{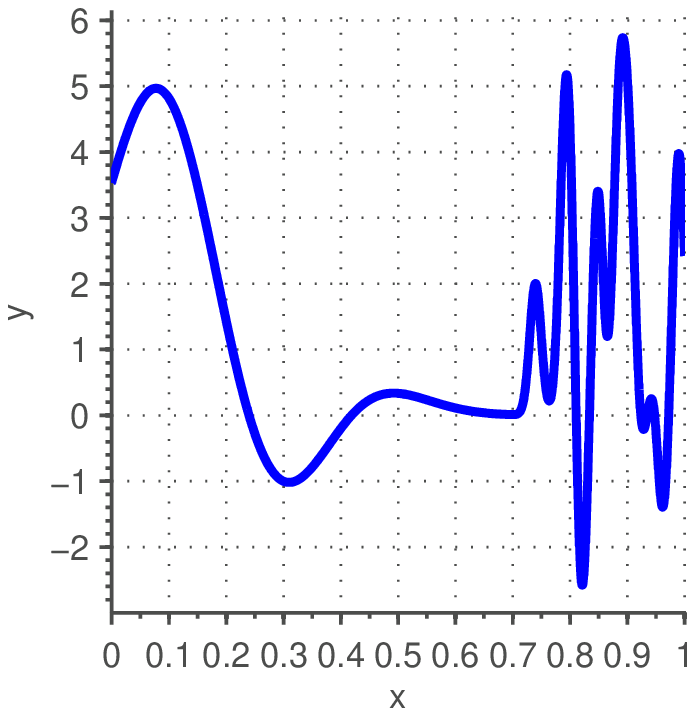}
	\includegraphics[width=0.48\linewidth]{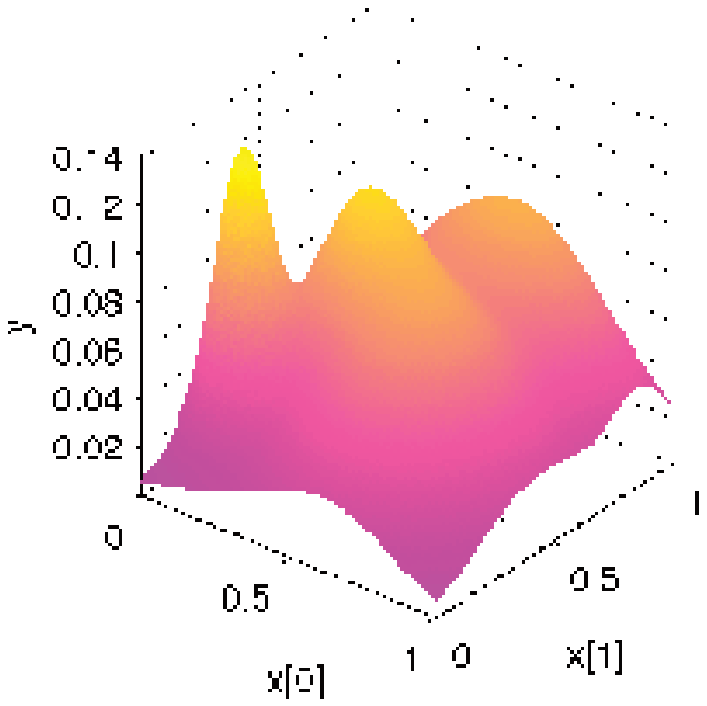}
    \caption{Left: RKHS function \cite{wang2014b}. Right: Gaussian mixture (GM) function}
    \label{fig:rkhs_function}
\end{figure}

In this section, we use two synthetic functions with distinct regions in terms of risk: the 1D RKHS from \cite{wang2014b} and a Mixture of 2D Gaussian distributions (GM), see Fig.\ref{fig:rkhs_function}. Both have a global maximum at a narrow peak, which represents a region of high risk, that may not be the ideal one in case of significant input noise. 

%%%%%%%%%%%%%%%%%%%%%%%%%%%%%%%%%% FIGURE %%%%%%%%%%%%%%%%%%%%%%%%%%%%%%%%%%
%%%%%%%%%%%%%%%%%%%%%%%%%%%%%%%%%% FIGURE %%%%%%%%%%%%%%%%%%%%%%%%%%%%%%%%%%

We have performed 100 runs of Bayesian Optimization for both functions (RKHS, GM) and the optimization procedure (BO and UBO). Also, to represent the input noise, we have used 100 Monte Carlo samples at the optimal query point as seen before. 

\begin{figure}[!htbp]
    \subfloat[$\bar{y}_{mc}\left(\mathbf{x}^*\right)$]{\includegraphics[width=.5\linewidth]{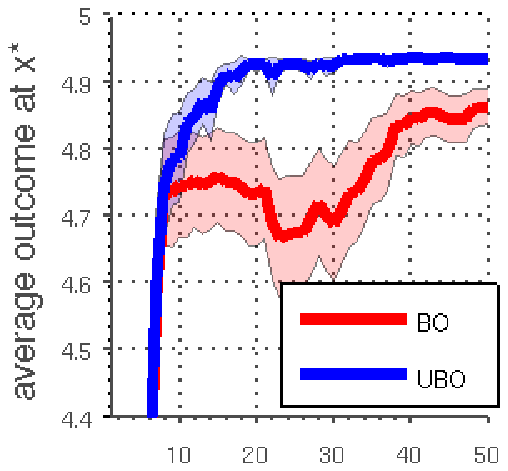}} \hfill
    %\subfloat[worst $y_{mc}\left(\mathbf{x}^*\right)$]{\includegraphics[width=.33\linewidth]{images/cuei_synth/rkhs_yworst.eps}} \hfill
    \subfloat[$std \left( y_{mc} \left(\mathbf{x}^*\right) \right)$]{\includegraphics[width=.5\linewidth]{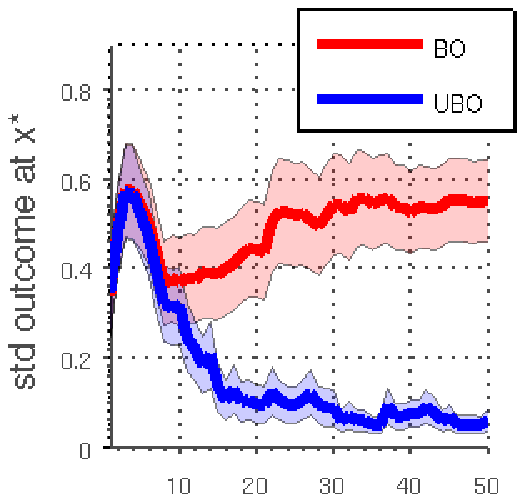}} \hfill
    \caption{RKHS Results}
    \label{fig:rkhs_cuei}
\end{figure}

\begin{figure}[!htbp]
    \subfloat[$\bar{y}_{mc}\left(\mathbf{x}^*\right)$]{\includegraphics[width=.5\linewidth]{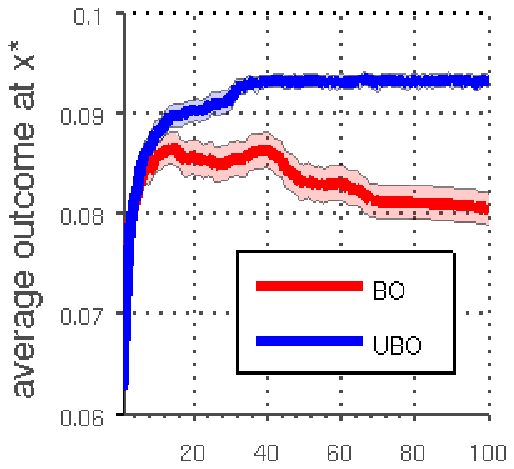}} \hfill
    %\subfloat[worst $y_{mc}\left(\mathbf{x}^*\right)$]{\includegraphics[width=.33\linewidth]{images/cuei_synth/martinez_yworst.eps}} \hfill
    \subfloat[$std \left( y_{mc} \left(\mathbf{x}^*\right) \right)$]{\includegraphics[width=.5\linewidth]{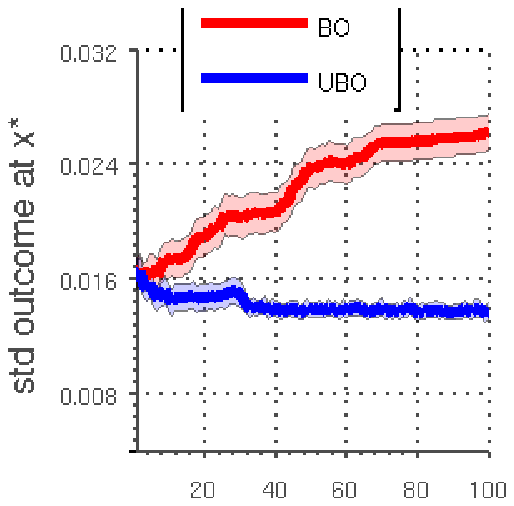}} \hfill
    \caption{G-Mixture Results}
    \label{fig:martinez_cuei}
\end{figure}

For RKHS each run has 5 initial samples and the optimization performs 45 iterations. The input noise is set as $\sigma_x = 0.01$. For GM each run has 30 initial samples and the optimization performs 90 iterations. The input noise is set as $\sigma_x = 0.1$. 
In Fig. \ref{fig:rkhs_cuei} and Fig. \ref{fig:martinez_cuei} we show the statistics over the different runs for the evaluation criteria with respect to the number of iterations. The shaded region represents the $95\%$ confidence interval. 

%For each criterion it was performed a total of 100 tests. For the RKHS function, each test had 5 initial random samples with a total budget of 50. While the Gaussian-Mixture function had 30 initial random samples and a total budget of 120. It was used $\sigma_y = 10^{-6}$ for both functions and . In each iteration it was taken 100 Monte Carlo samples.

%%%%%%%%%%%%%%%%%%%%%%%%%%%%%%%%%% FIGURE %%%%%%%%%%%%%%%%%%%%%%%%%%%%%%%%%%
 %%%%%%%%%%%%%%%%%%%%%%%%%%%%%%%%%% FIGURE %%%%%%%%%%%%%%%%%%%%%%%%%%%%%%%%%%
 
%%%%%%%%%%%%%%%%%%%%%%%%%%%%%%%%%% FIGURE %%%%%%%%%%%%%%%%%%%%%%%%%%%%%%%%%%
 %%%%%%%%%%%%%%%%%%%%%%%%%%%%%%%%%% FIGURE %%%%%%%%%%%%%%%%%%%%%%%%%%%%%%%%%%

For both functions, we can observe that UBO quickly overcomes the results of BO. As soon as the random exploration phase finishes and the optimization starts, the UBO computes less risky solutions, as demonstrated by the higher expected return value and lower standard deviation. In table \ref{tbl:results} we show the numeric results obtained at the last iteration. We show also values for the worst sample of the Monte Carlo runs\footnote{Worst cases are not shown graphically due to lack of space, but they are coherent with the evolution of the means.}. The worst case for UBO is always more favourable than the worst case for BO by a large margin.

%better expected return values (assess via the mean of the Monte Carlo samples) chooses a high-valued optimum in terms of safety. This is ostended by having higher mean output inside a vicinity of the optimum than the expected improvement. The other two measures of unsafetiness - worst ouput and standard deviation of the outputs - are respectively higher and lower for the UBO which also confirms the previous observation. All results can be inspected in table \ref{tbl:results}.

    \subsection{Robot Grasp Simulations}
    
%Bayesian Optimization will be performed for a simulated learning task - find the optimal grasp for a certain object. 

\begin{figure}[!htb]
	\centering
	\includegraphics[width=0.70\linewidth]{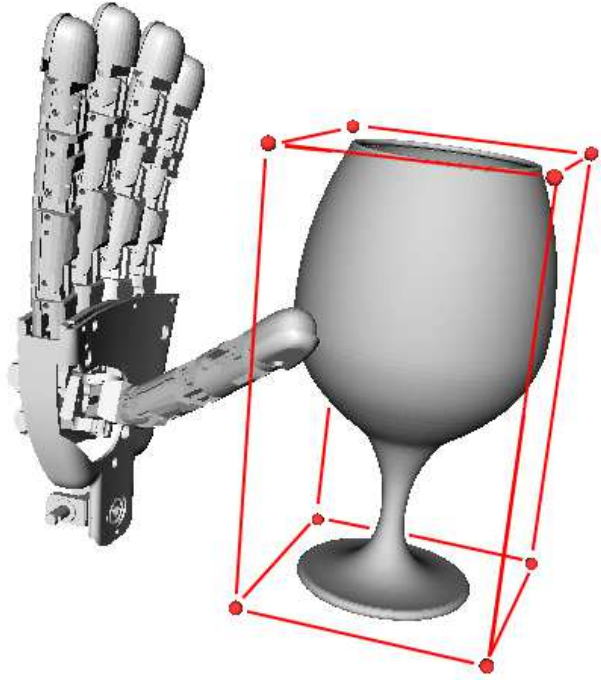}
    \caption{The Simox robot grasp simulator. The iCub's left hand is used to perform grasping trials on arbitrary objects, in this case a glass. The red lines around the glass represent and Object Oriented Bounding Box, whose facets are used to setup the initial hand configuration.}
    \label{fig:obb}
\end{figure}

\begin{figure}[!htbp]
    \subfloat[Waterbottle]{\includegraphics[width=.45\linewidth]{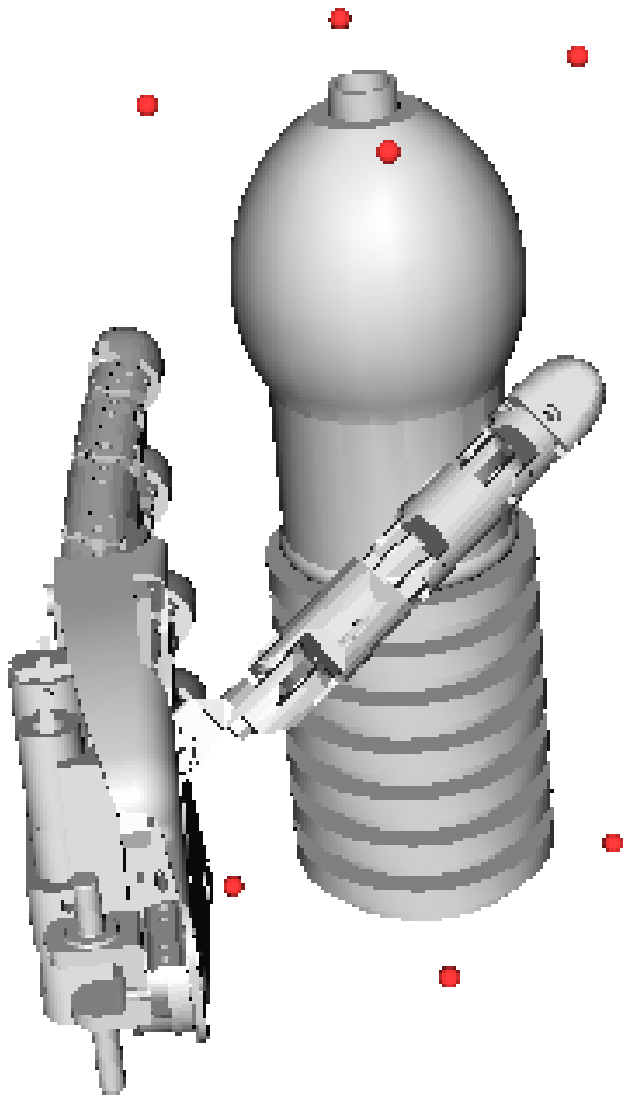}} \hfill
    \subfloat[Mug]{\includegraphics[width=.45\linewidth]{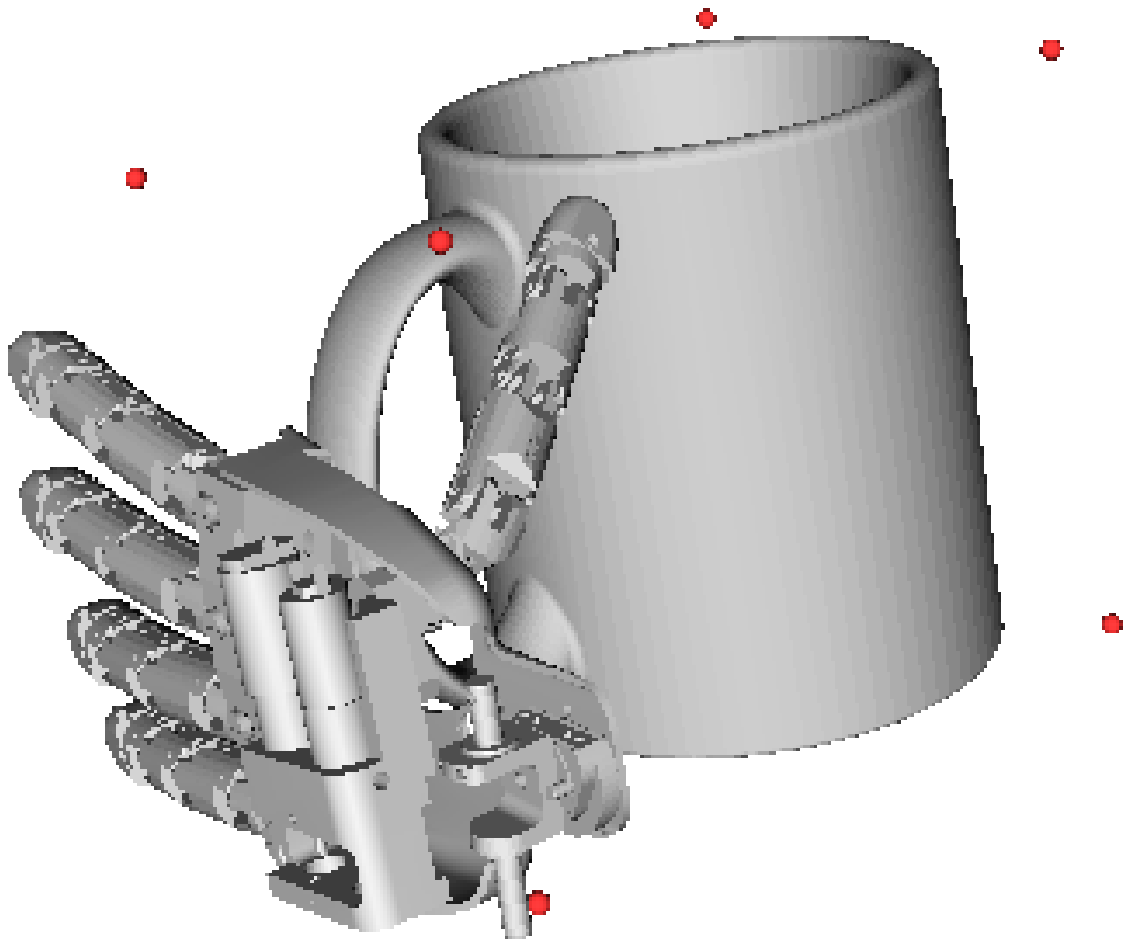}} \hfill
    \subfloat[Glass]{\includegraphics[width=.45\linewidth]{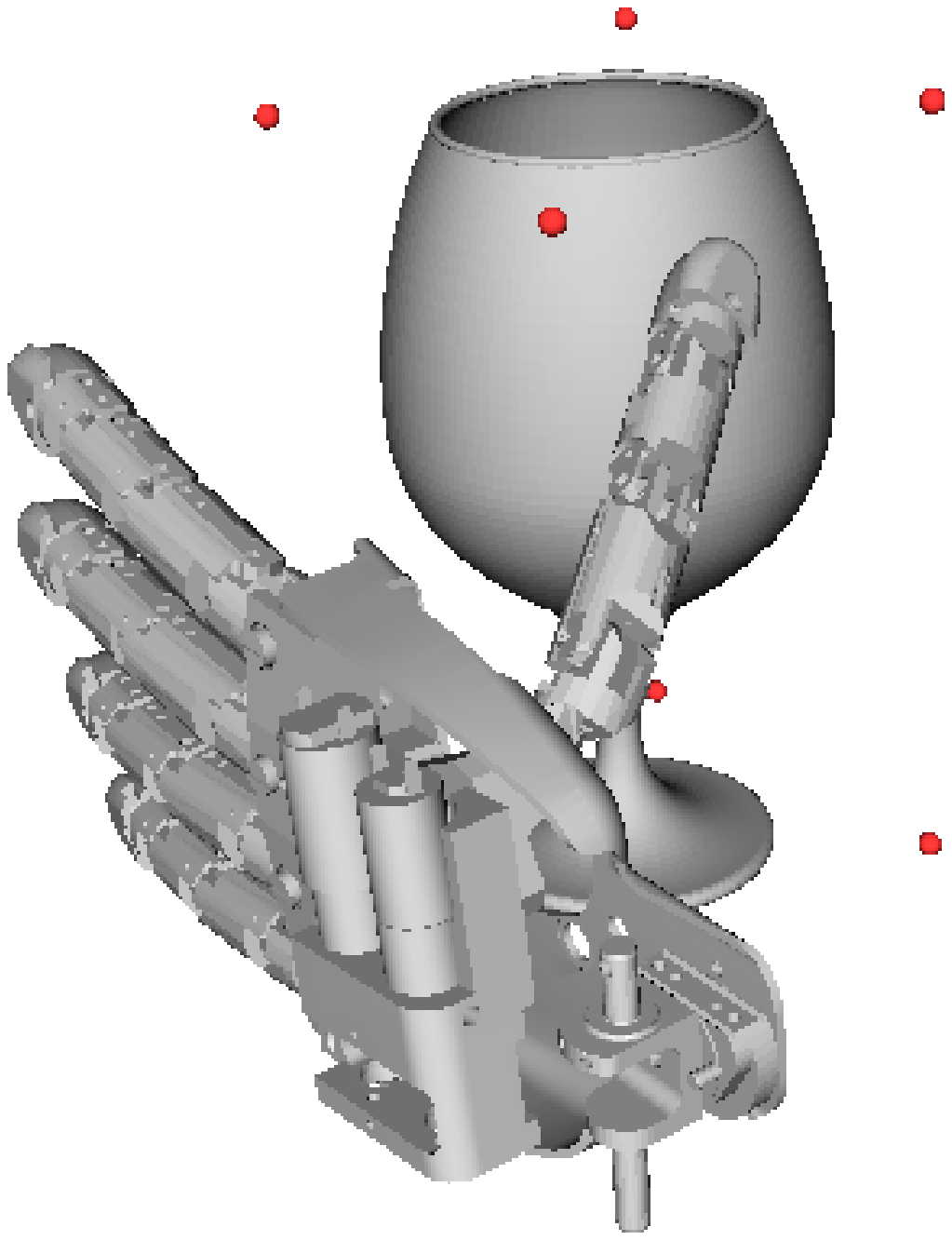}} \hfill
    \subfloat[Drill]{\includegraphics[width=.45\linewidth]{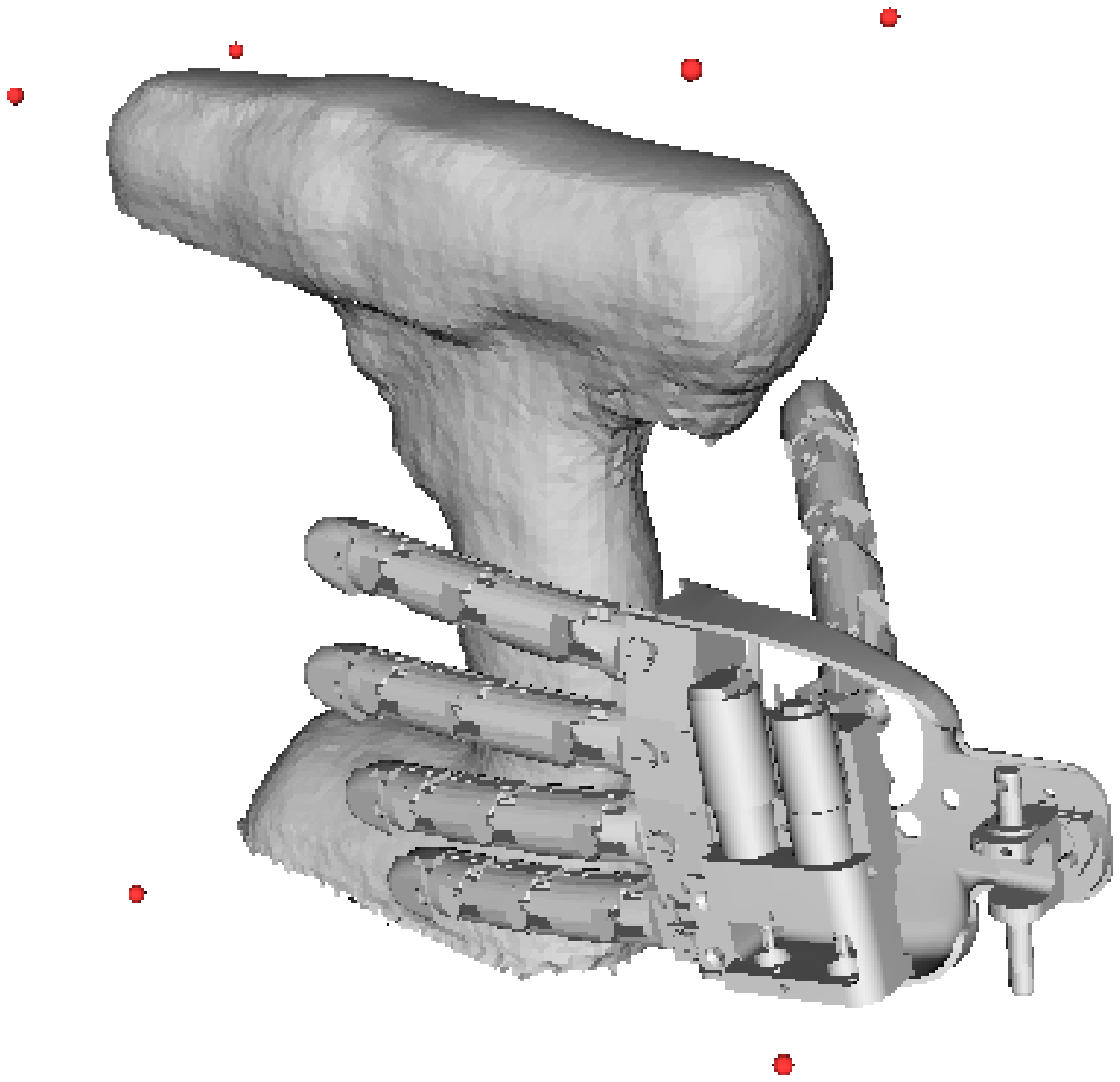}} \hfill

    \caption{Objects used in the simulations with corresponding initial robot hand configuration.}
    \label{fig:objects}
\end{figure}

We use the Simox simulation toolbox for robot grasping \cite{simox}. This toolbox simulates the iCub's robot hand grasping arbitrary objects. Given an initial pose for the robot hand and a finger joint trajectory, the simulator runs until the fingers are in contact with the object surface and computes a grasp quality metric based on wrench space analysis. We use a representation of the iCub's left hand which can move freely in space (Fig. \ref{fig:obb}) and a few static objects, shown in \ref{fig:objects}.

The robot hand is initially placed with palm facing parallel to one of the facets, at a fixed distance of the object bounding box, and the thumb aligned with one of the neighbour facets (Fig. \ref{fig:obb}). This defines uniquely the pose of the hand with respect to the object. The hand's pose is then defined with respect to the initial pose by incremental translations and rotations: $ \left( \delta_x, \delta_y, \delta_z, \theta_x, \theta_y, \theta_z \right)$. 

For the experiments we have only optimized the translation parallel to the facet $ \left( \delta_x, \delta_y\right)$  while the rest of the values were fixed in advance. The translation variables are bound to the limits of the bounding box. A power grasp posture synergy was adopted for the hand closure. In this strategy all finger joints close at the same time by roughly the same amount. The advantage of the Bayesian optimization methodology as a black-box optimization is that the system is agnostic to the parametrization selected and it can be easily replaced.

%To learn the optimal grasp, a search over the relative position between hand and object and also hand closure primitives. Each experiment is performed for one type of grasp and relative to one of six possible facets of the object \cite{babbling2009} - obtained through an object oriented bounding box (Fig. \ref{fig:obb}). All objects used can be seen in Fig. \ref{fig:objects}

%%%%%%%%%%%%%%%%%%%%%%%%%%%%%%%%%% FIGURE %%%%%%%%%%%%%%%%%%%%%%%%%%%%%%%%%%
%%%%%%%%%%%%%%%%%%%%%%%%%%%%%%%%%% FIGURE %%%%%%%%%%%%%%%%%%%%%%%%%%%%%%%%%%

%%%%%%%%%%%%%%%%%%%%%%%%%%%%%%%%%% FIGURE %%%%%%%%%%%%%%%%%%%%%%%%%%%%%%%%%%

We have performed 30 runs of the robotic grasp simulation for each object and each optimization criterion. 
The robot hand posture with respect to the objects is initialized as shown in Fig. \ref{fig:objects}.

\begin{figure}[!htbp]
    \subfloat[$\bar{y}_{mc}\left(\mathbf{x}^*\right)$]{\includegraphics[width=.5\linewidth]{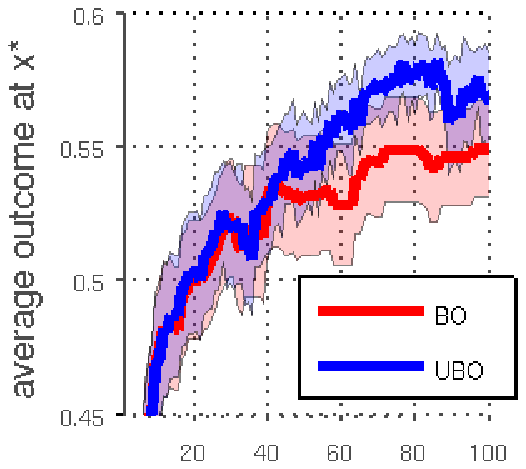}} \hfill
    %\subfloat[worst $y_{mc}\left(\mathbf{x}^*\right)$]{\includegraphics[width=.33\linewidth]{images/cuei_simox/waterbottle_yworst.eps}} \hfill
    \subfloat[$std \left( y_{mc} \left(\mathbf{x}^*\right) \right)$]{\includegraphics[width=.5\linewidth]{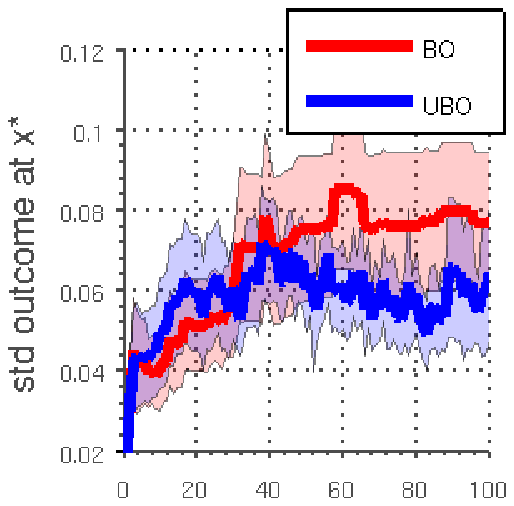}} \hfill
    \caption{Waterbottle.
    %iCub - Waterbottle - RI-FR -  PowerGrip - $ \sigma_y = 1e-4 $ - 
    Input Space Noise $\sigma_x = 0.01$}
    \label{fig:waterbottle_cuei}
\end{figure}

\begin{figure}[!htbp]
    \subfloat[$\bar{y}_{mc}\left(\mathbf{x}^*\right)$]{\includegraphics[width=.5\linewidth]{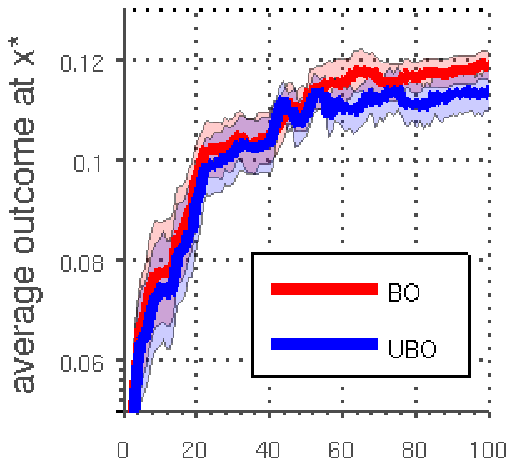}} \hfill
    %\subfloat[worst $y_{mc}\left(\mathbf{x}^*\right)$]{\includegraphics[width=.33\linewidth]{images/cuei_simox/mug_yworst.eps}} \hfill
    \subfloat[$std \left( y_{mc} \left(\mathbf{x}^*\right) \right)$]{\includegraphics[width=.5\linewidth]{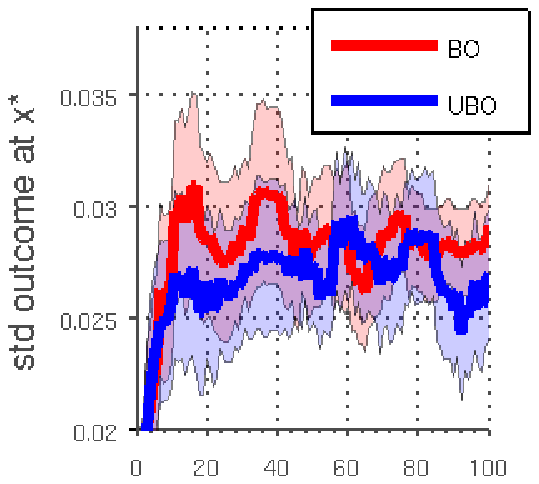}} \hfill
    \caption{Mug.
    %iCub - Mug - LE-BA - PowerGrip - $ \sigma_y = 1e-4 $ - 
    Input Space Noise $\sigma_x = 0.03$}
    \label{fig:mug_cuei}
\end{figure}

\begin{figure}[!htbp]
    \subfloat[$\bar{y}_{mc}\left(\mathbf{x}^*\right)$]{\includegraphics[width=.5\linewidth]{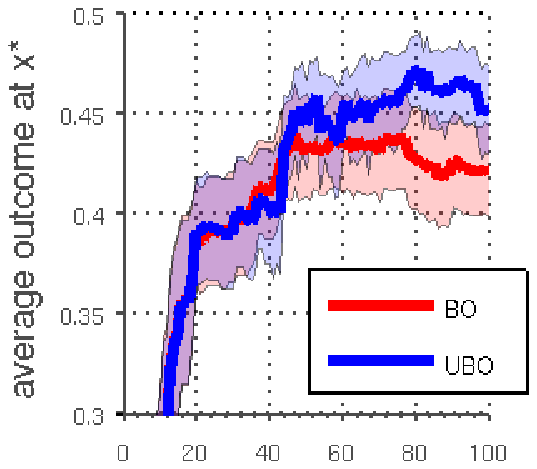}} \hfill
    %\subfloat[worst $y_{mc}\left(\mathbf{x}^*\right)$]{\includegraphics[width=.33\linewidth]{images/cuei_simox/glass_yworst.eps}} \hfill
    \subfloat[$std \left( y_{mc} \left(\mathbf{x}^*\right) \right)$]{\includegraphics[width=.5\linewidth]{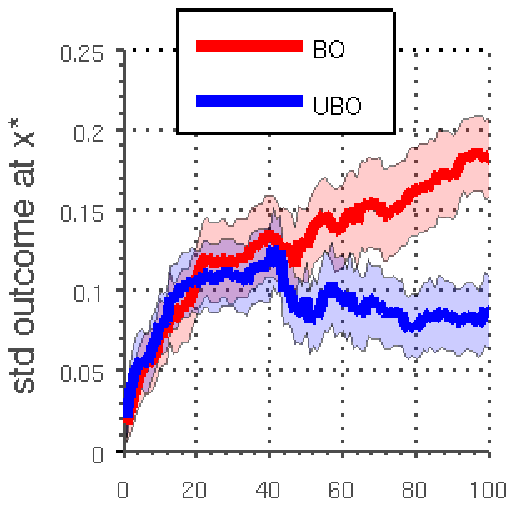}} \hfill
    \caption{Glass.
    %iCub - Glass - RI-FR - PowerGrip - $ \sigma_y = 1e-4 $ - 
    Input Space Noise $\sigma_x = 0.03$}
    \label{fig:glass_cuei}
\end{figure}

\begin{figure}[!htbp]
    \subfloat[$\bar{y}_{mc}\left(\mathbf{x}^*\right)$]{\includegraphics[width=.5\linewidth]{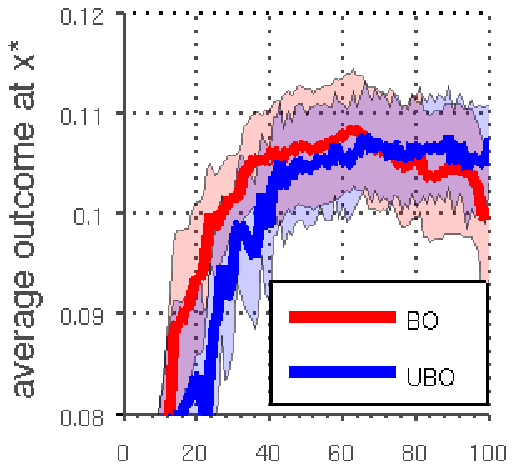}} \hfill
    %\subfloat[worst $y_{mc}\left(\mathbf{x}^*\right)$]{\includegraphics[width=.33\linewidth]{images/cuei_simox/drill_yworst.eps}} \hfill
    \subfloat[$std \left( y_{mc} \left(\mathbf{x}^*\right) \right)$]{\includegraphics[width=.5\linewidth]{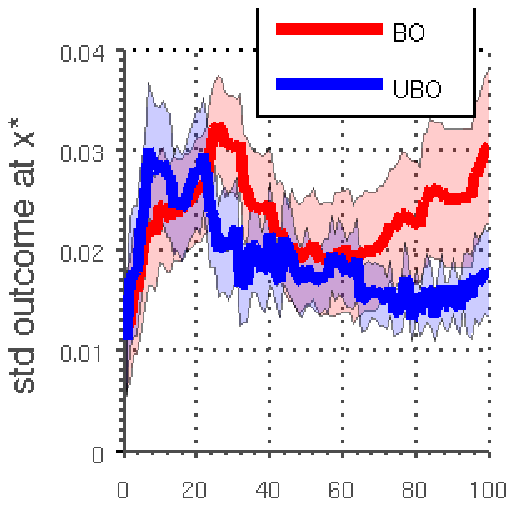}} \hfill
    \caption{Drill.
    %iCub - Drill - RI-FR - PowerGrip - $ \sigma_y = 1e-4 $ - 
    Input Space Noise $\sigma_x = 0.03$}
    \label{fig:drill_cuei}
\end{figure}

Each run starts with 40 initial random samples and proceeds with 60 iterations of Bayesian optimization, for a total of 100. In this case, we assume that the function is stochastic, due to small simulation errors and inconsistencies, assuming $\sigma_y = 10^{-4}$. Also, we assume an input noise $\sigma_x = 0.03$ (note that the input space was normalized in advance to the unit hypercube $[0,1]^d$). In each iteration we sample 20 times at the query point with input noise to compute the expected outcome. The results can be observed in figures \ref{fig:waterbottle_cuei}, \ref{fig:mug_cuei}, \ref{fig:glass_cuei} and \ref{fig:drill_cuei}. We note that the plots seem noisier than with the synthetic functions. This fact is due to a lower number of samples at the query points, for the sake of computation time, as each one required to run the full grasp simulation. Note also that those samples are only used for illustrative purposes and would not be required in practice. 

%%%%%%%%%%%%%%%%%%%%%%%%%%%%%%%%%% FIGURE %%%%%%%%%%%%%%%%%%%%%%%%%%%%%%%%%%
 %%%%%%%%%%%%%%%%%%%%%%%%%%%%%%%%%% FIGURE %%%%%%%%%%%%%%%%%%%%%%%%%%%%%%%%%%
 
%%%%%%%%%%%%%%%%%%%%%%%%%%%%%%%%%% FIGURE %%%%%%%%%%%%%%%%%%%%%%%%%%%%%%%%%%
 %%%%%%%%%%%%%%%%%%%%%%%%%%%%%%%%%% FIGURE %%%%%%%%%%%%%%%%%%%%%%%%%%%%%%%%%%
 
 %%%%%%%%%%%%%%%%%%%%%%%%%%%%%%%%%% FIGURE  %%%%%%%%%%%%%%%%%%%%%%%%%%%%%%%%%%
 
%%%%%%%%%%%%%%%%%%%%%%%%%%%%%%%%%% FIGURE %%%%%%%%%%%%%%%%%%%%%%%%%%%%%%%%%%
 %%%%%%%%%%%%%%%%%%%%%%%%%%%%%%%%%% FIGURE %%%%%%%%%%%%%%%%%%%%%%%%%%%%%%%%%%
 
 %%%%%%%%%%%%%%%%%%%%%%%%%%%%%%%%%% FIGURE %%%%%%%%%%%%%%%%%%%%%%%%%%%%%%%%%%
 %%%%%%%%%%%%%%%%%%%%%%%%%%%%%%%%%% FIGURE %%%%%%%%%%%%%%%%%%%%%%%%%%%%%%%%%%
 
It can be observed that, for the water bottle and glass, the UBO method has clear advantages over BO. As soon as the initial sampling phase finishes, UBO obtains higher mean values and lower standard deviations. For the drill, the UBO eventually overcomes the BO, but at later iterations, which might imply that the unsafe optimum is difficult to find, but still exists. Looking at the quantitative results shown in Table \ref{tbl:results}, we can see that, at the end of the optimization, UBO is better than BO in all criteria, except for the mean output value for the mug. For the mug, the 100 iteration are not enough to obtain better mean values. We can see that the mug and drill objects are more challenging due to their non-rotational symmetry. Since the optimization is only done in translation parameters, the method is missing exploration in the rotation degrees of freedom. Furthermore, in the mug's case, the facet chosen was the one that contains the mug's handle. Trying to learn a grasp in this setting is much harder than the other cases since, for the same input space volume, the percentage of configurations which return a good metric is much smaller. This hinders Bayesian Optimization performance in general and deteriorates GP regression. For the water bottle and the glass, the rotational degrees of freedom are not so important because the objects are rotationally symmetric. 

%These experiments have the must relevant results of this work, since we couldn't not infer directly if the UBO would perform better than normal BO in terms of risk in the grasp learning setting. This research was lead by the observation that not all grasps are optimal, even if they have high output metrics. Optimal grasps must also need to be stable in terms of noisy controls (which is not modeled in those metrics). 

%It can be examined that UBO performs better than normal BO for the overall set of objects. The best results were obtained for the water bottle and glass cup, in which the UBO distinguish the most from BO. Only the mug has low results for the UBO - in terms of the Monte Carlo sample outputs' mean. This is most probably due to the fact that the facet chosen was the one that contains the mug's handle. Trying to learn a grasp in this setting is much harder than the other cases since, for the same input space volume, the percentage of configurations which return a good metric is much smaller. This hinders Bayesian Optimization performance in general and deteriorates GP regression.

\begin{figure}[!htbp]
    \subfloat[Safe-zone, $y = 0.413$]{\includegraphics[width=.45\linewidth]{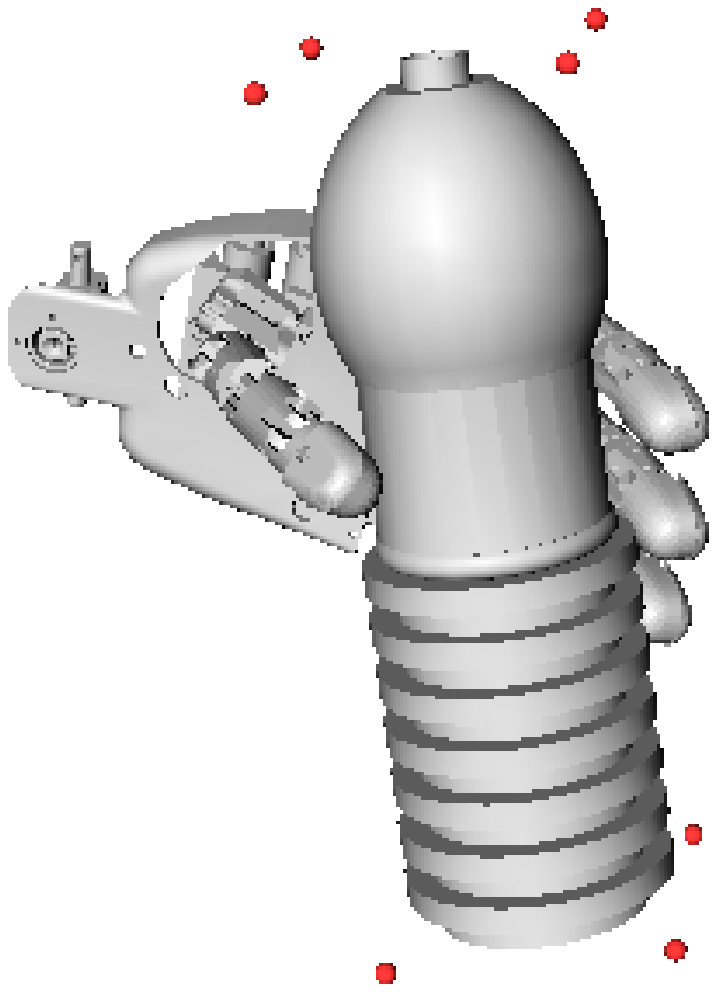}} \hfill
    \subfloat[Safe-zone, $y = 0.418$]{\includegraphics[width=.45\linewidth]{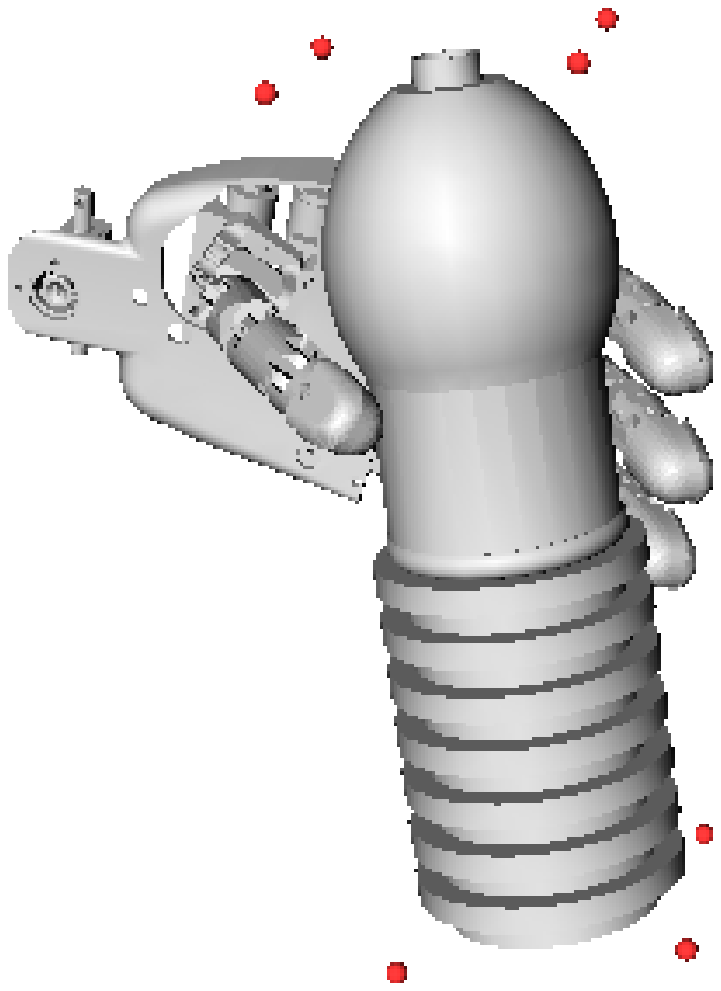}} \hfill
    \subfloat[Unsafe-zone, $y = 0.439$]{\includegraphics[width=.45\linewidth]{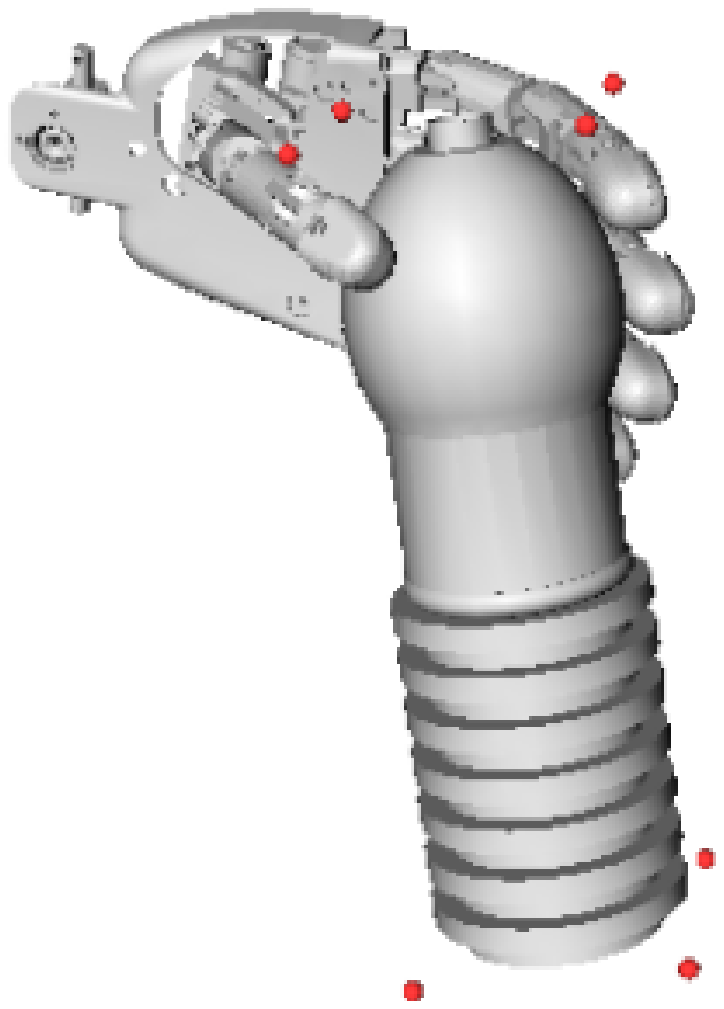}} \hfill
    \subfloat[Unsafe-zone, $y = 0.377$]{\includegraphics[width=.45\linewidth]{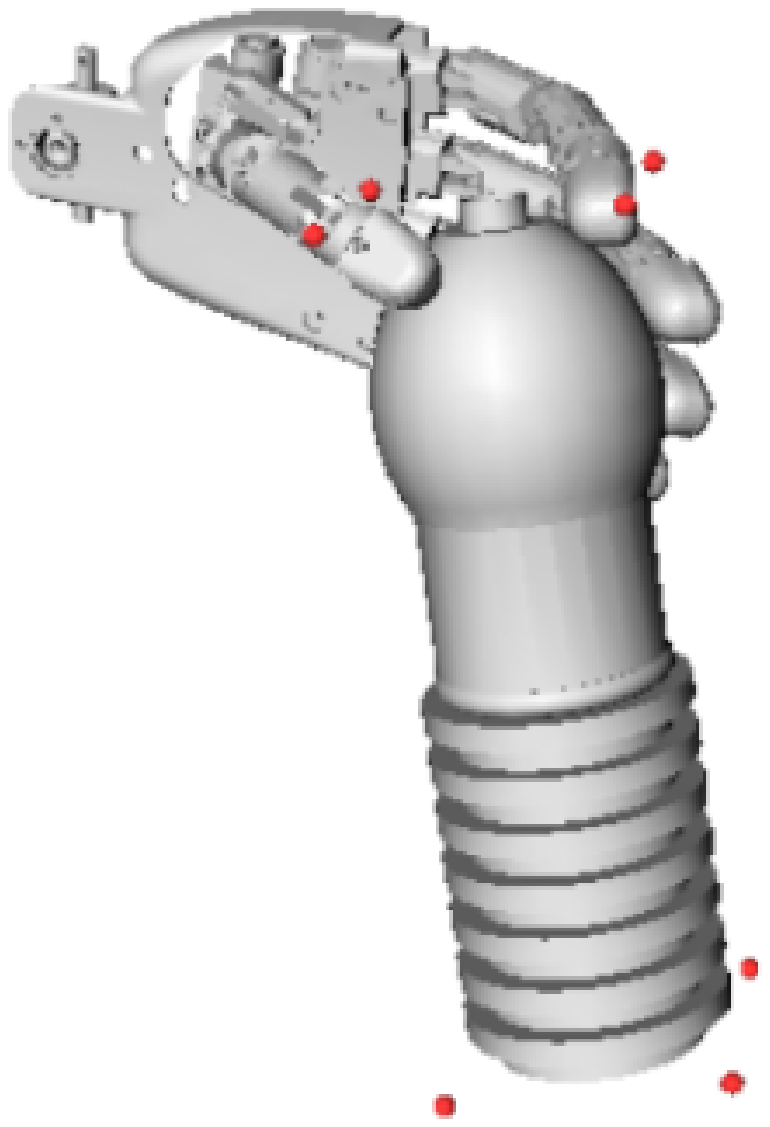}} \hfill

    \caption{Grasp safety. In this example the best grasp is at an unsafe zone (c). However, a bad grasp is in its vicinity (d). The unscented Bayesian optimization chooses grasps with lower risk at the safe zone (a) and (b), where performance is robust to input noise. }
    \label{fig:safeness}
\end{figure}

In Fig. \ref{fig:safeness} we illustrate four grasps at the water bottle explored during the experiments. Two of the grasps are performed in a safe region while the two other are explored at a unsafe region. Although the unsafe zone has one observation with the highest value, it has also higher risk of getting a low value observation in its vicinity. 

%%%%%%%%%%%%%%%%%%%%%%%%%%%%%%%%%% FIGURE %%%%%%%%%%%%%%%%%%%%%%%%%%%%%%%%%%
 %%%%%%%%%%%%%%%%%%%%%%%%%%%%%%%%%% FIGURE %%%%%%%%%%%%%%%%%%%%%%%%%%%%%%%%%%

 %\subsection{Experiments - Simulated Environment Application}
 
 %%%%%%%%%%%%%%%%%%%%%%%%%%%%%%%%%% TABLE %%%%%%%%%%%%%%%%%%%%%%%%%%%%%%%%%%

\begin{table}[!htb]
    \centering
    \begin{supertabular}[c]{| p{1cm} | p{0,7cm} | p{0,7cm} | p{0,7cm} | p{0,7cm} | p{0,7cm} | p{0,7cm} |}
    
        \hline
        \multicolumn{1}{| c |}{}                                             &
        \multicolumn{2}{ c |}{$\bar{y}_{mc}\left(\mathbf{x}^*\right)$}       &
        \multicolumn{2}{ c |}{$worst \quad y_{mc}\left(\mathbf{x}^*\right)$} &
        \multicolumn{2}{ c |}{$std \left( y_{mc} \left(\mathbf{x}^*\right) \right)$}                                                            \\
        
        \hline
        \multicolumn{7}{ c }{Synthetic Problems} \\
        
        \hline
        Function &
        BO       &
        UBO      &
        BO       &
        UBO      &
        BO       &
        UBO      \\
        
        \hline
        RKHS           &
        $    4.863 $   &
        $\bf{4.934}$   &
        $    2.881 $   &
        $\bf{4.657}$   &
        $    0.554 $   &
        $\bf{0.065}$   \\

        \hline
        G-Mix          &
        $    0.080 $   &
        $\bf{0.093}$   &
        $    0.023 $   &
        $\bf{0.053}$   &
        $    0.027 $   &
        $\bf{0.014}$   \\
        
        \hline
        \multicolumn{7}{ c }{Simulation - Simox} \\
        
        \hline
        Object   &
        BO       &
        UBO      &
        BO       &
        UBO      &
        BO       &
        UBO      \\
        
        \hline
        Bottle    &
        $    0.550 $   &
        $\bf{0.567}$   &
        $    0.390 $   &
        $\bf{0.430}$   &
        $    0.077 $   &
        $\bf{0.065}$   \\

        \hline
        Mug            &
        $\bf{0.119}$   &
        $    0.114 $   &
        $    0.051 $   &
        $\bf{0.059}$   &
        $    0.029 $   &
        $\bf{0.027}$   \\
        
        \hline
        Glass          &
        $    0.421 $   &
        $\bf{0.452}$   &
        $    0.080 $   &
        $\bf{0.252}$   &
        $    0.184 $   &
        $\bf{0.087}$   \\
        
        \hline
        Drill          &
        $    0.101 $   &
        $\bf{0.108}$   &
        $    0.050 $   &
        $\bf{0.068}$  &
        $    0.030 $   &
        $\bf{0.018}$   \\
        \hline
        
     \end{supertabular}
     
    \caption{Results at the last iteration of the Bayesian Optimization process (means over all runs).}
    \label{tbl:results}
\end{table}

%%%%%%%%%%%%%%%%%%%%%%%%%%%%%%%%%% TABLE %%%%%%%%%%%%%%%%%%%%%%%%%%%%%%%%%%

\section{Conclusion} 
\label{sec:conclusion}

The contribution of this paper is twofold. On one hand, we present a method for \emph{robust and safe grasping} of unknown objects by trial and error, analogous to the process infants develop object exploration and grasping at early stages. Because the process is based on general black-box optimization, the method is capable to deal with arbitrary objects, effectors and environmental conditions. On the other hand, in this work we have developed an extension for \emph{Bayesian optimization in the presence of input noise}, that we have called unscented Bayesian optimization. The potential interest of this method goes beyond grasping or even robotics. Bayesian optimization is currently being used in many applications, from engineering, computer sciences, economics, simulations, experimental design, biology, artificial intelligence, etc. In all those fields, there are many situations where input noise or uncertainty may arise. In those cases, safe optimization is fundamental. For example, in other areas of robotics, it might be used for navigation, planing or sensor placement, as the robot/sensor location is uncertain.

%\section*{Acknowledgments}

%% Use plainnat to work nicely with natbib. 

\bibliographystyle{plain}
\bibliography{references}

\end{document}